\documentclass{article}
\usepackage[preprint]{spconf,amsmath,graphicx,amssymb}
\copyrightnotice{\copyright\ IEEE 2019}
\toappear{To appear in {\it Proc.\ ASRU 2019, December 14-18, 2019, Sentosa, Singapore}}
\usepackage{amsfonts}
\usepackage{xspace}
\usepackage[smallerops]{newtxmath}
\usepackage{bm}
\usepackage{csquotes}
\usepackage{todonotes}
\usepackage{nicefrac}
\usepackage{hyperref}
\usepackage{listings}
\usepackage{booktabs}
\usepackage{tablefootnote}
\usepackage{bbm}
\usepackage{multirow}
\usepackage{multicol}
\usepackage{adjustbox}
\usepackage{xcolor}
\usepackage{float}
\usepackage{accents}
\usepackage{enumitem}
\renewcommand*\ttdefault{txtt}

\def\espresso{\textsc{Espresso}\xspace}
\def\fairseq{\textsc{fairseq}\xspace}
\def\opennmt{\textsc{OpenNMT}\xspace}
\def\openseqseq{\textsc{OpenSeq2Seq}\xspace}
\def\espnet{\textsc{ESPnet}\xspace}
\def\eesen{\textsc{EESEN}\xspace}
\def\lingvo{\textsc{lingvo}\xspace}
\def\returnn{\textsc{RETURNN}\xspace}
\def\wavtoletterpp{\textsc{wav2letter++}\xspace}

\DeclareMathAlphabet{\mathsf}{OT1}{\sfdefault}{m}{n}
\DeclareMathAlphabet{\mathtt}{OT1}{\ttdefault}{m}{n}
\SetMathAlphabet{\mathtt}{bold}{OT1}{\ttdefault}{b}{n}

\def\prefix{\textit{p}}
\def\prefixq{\textit{q}}
\def\suffix{\textit{s}}
\def\word{{\textit{w}}}
\def\ch{\mathtt{c}}
\def\history{\mathbf{h}}

\setlist[itemize]{leftmargin=0.5cm}

\title{\espresso: A Fast End-to-end Neural Speech Recognition Toolkit}
\name{\parbox{\linewidth}{\centering
Yiming Wang$^1$, Tongfei Chen$^1$, Hainan Xu$^1$, Shuoyang Ding$^1$, Hang Lv$^{1,4}$, Yiwen Shao$^1$, \\ \textit{Nanyun Peng$^3$, Lei Xie$^4$, Shinji Watanabe$^1$, Sanjeev Khudanpur$^{1,2}$}\thanks{This work was partially supported by the \href{https://www.iarpa.gov/index.php/research-programs/material}{IARPA MATERIAL} program and by an unrestricted gift from \href{https://www.mobvoi.com/us/pages/about-us}{Mobvoi}. The authors also thank Ziyan Jiang for running speed tests on \espnet.}}}

\address{\textsuperscript{1}~Center of Language and Speech Processing, \textsuperscript{2}~Human Language Technology Center of Excellence, \\ Johns Hopkins University, Baltimore, MD, USA \\ \textsuperscript{3}~Information Sciences Institute, University of Southern California, Los Angeles, CA, USA \\ \textsuperscript{4}~ASLP@NPU, School of Computer Science, Northwestern Polytechnical University, Xi'an, China \\ \footnotesize{\texttt{\string{yiming.wang,tongfei,hxu31,dings,yshao18,shinjiw,khudanpur\string}@jhu.edu, npeng@isi.edu, \string{hanglv,lxie\string}@nwpu-aslp.org}}}

\begin{document}
\ninept

\maketitle
\begin{abstract}
  We present \espresso, an open-source, modular, extensible end-to-end neural automatic speech recognition (ASR) toolkit based on the deep learning library PyTorch and the popular neural machine translation toolkit \fairseq. \espresso supports distributed training across GPUs and computing nodes, and features various decoding approaches commonly employed in ASR, including look-ahead word-based language model fusion, for which a fast, parallelized decoder is implemented. 
  \espresso achieves state-of-the-art ASR performance on the WSJ, LibriSpeech, and Switchboard data sets among other end-to-end systems without data augmentation, and is 4--11$\times$ faster for decoding than similar systems (e.g. \espnet).
\end{abstract}

\begin{keywords}
automatic speech recognition, end-to-end, parallel decoding, language model fusion
\end{keywords}

\section{Introduction}
\label{sec:intro}
Various open-source toolkits for building automatic speech recognition (ASR) systems have been created, with a notable example being Kaldi \cite{povey2011kaldi}, a weighted finite state transducer based framework with extensive linear algebra support, that enables traditional hybrid ASR systems \cite{yu2014automatic}.

With advances in deep learning, recent work in ASR begun paying attention to so-called \emph{neural end-to-end} systems \cite[\emph{inter alia}]{graves2014towards,chorowski2014end,chan2016listen}, which are characterized by generally smaller code size, and greater portability and maintainability across hardware platforms and software environments. This shift is analogous to the one in the machine translation (MT) community: from feature- and syntax-based statistical machine translation (SMT) systems (e.g. Moses \cite{koehn2007moses}, Joshua \cite{li2009joshua}) to end-to-end neural machine translation (NMT) systems (e.g. \opennmt \cite{klein2017opennmt}, \openseqseq \cite{openseq2seq}, \fairseq \cite{ott2019fairseq}). As a result, there have been a few efforts in the ASR research community to create open source neural end-to-end  frameworks, most notably \espnet \cite{watanabe2018espnet} (See also, Table~\ref{tab:e2e_systems}). However, \espnet has some important shortcomings: (i) the code is not very easily extensible and has portability issues due to its mixed dependency on two deep learning frameworks PyTorch \cite{paszke2017automatic} and Chainer \cite{seiya2015chainer}; (ii) the decoder, which uses a simple but relatively slow beam search algorithm, is not fast enough for quick turnaround of experiments.

\begin{table*}[t!]
  \caption{Popular end-to-end neural ASR systems and our system.}
  \label{tab:e2e_systems}
  \centering
  \begin{adjustbox}{max width=\textwidth}
  \begin{tabular}{l c c c c}
    \toprule
    Name & Language & Deep Learning Framework & Recipes & Other Applications\\
    \midrule
    \eesen \cite{miao2015end} & C++ & --- & WSJ, LibriSpeech, SWBD, TED-LIUM, HKUST & ---\\
    \espnet \cite{watanabe2018espnet} & Python & Chainer \& PyTorch & various & TTS, ST\\
    E2E LF-MMI \cite{hadian2018end} & C++ & Kaldi & various & ---\\
    \lingvo \cite{shen2019lingvo} & Python & TensorFlow & LibriSpeech & NMT \\
    \openseqseq \cite{openseq2seq} & Python & TensorFlow & LibriSpeech & NMT, TTS \\
    \returnn \cite{zeyer2018returnn} & Python & Theano, TensorFlow & WSJ, LibriSpeech, SWBD, CHiME & NMT \\
    \wavtoletterpp \cite{pratap2019wav2letter} & C++ & ArrayFire & WSJ, LibriSpeech, TIMIT & ---\\
    \midrule
    \espresso & Python & PyTorch & WSJ, LibriSpeech, SWBD & NMT \\
    \bottomrule
  \end{tabular}
  \end{adjustbox}
\end{table*}
To address these problems, we present \espresso, a novel neural end-to-end system for ASR.\footnote{~\url{https://github.com/freewym/espresso}.} \espresso builds upon the popular NMT framework \fairseq\footnote{After our submission, \fairseq released their official ASR support in \url{https://github.com/pytorch/fairseq/tree/master/examples/speech_recognition}, and a Transformer-based LibriSpeech recipe \cite{mohamed2019transformers}.}, and the flexible deep learning framework PyTorch. By extending \fairseq, \espresso inherits its excellent extensibility: new modules can easily be plugged into the system by extending standard PyTorch interfaces. Additionally, we gain ability to perform distributed training over large data sets for ASR.

We also present the first fully parallelized decoder for end-to-end ASR, with look-ahead word-based language model fusion \cite{hori2018endtoend}, tightly integrated with the existing sets of optimized inference algorithms (e.g. beam search) inherited from \fairseq and tailored to the scenario of speech recognition. Furthermore, an improved {\em coverage} mechanism is proposed to further reduce deletion and insertion errors, and is compared with related techniques such as EOS threshold \cite{hannun2019sequence}. \espresso provides recipes for a variety of benchmark ASR data sets, including WSJ \cite{paulb92}, LibriSpeech \cite{panayotovcpk15}, and Switchboard \cite{godfrey1992switchboard}, and achieves \emph{state-of-the-art results} on these data sets.

\espresso, by building upon \fairseq, also has the potential to integrate seamlessly with sequence generation systems from natural language processing (NLP), such as neural machine translation and dialog systems. We envision that \espresso could become the foundation for unified speech + text processing systems, and pave the way for future end-to-end speech translation (ST) and text-to-speech synthesis (TTS) systems, ultimately facilitating greater synergy between the ASR and NLP research communities.

\section{Software Architecture and Design Choices}
\label{sec:design}

We implement \espresso with the following design goals in mind:
\begin{itemize} \setlength\itemsep{0cm}
    \item Pure Python / PyTorch that enables modularity and extensibility;
    \item Parallelization and distributed training and decoding for quick turnaround of experiments;
    \item Compatibility with Kaldi / \espnet data format to enable reuse of previous / proven data preparation pipelines;
    \item Easy interoperability with the existing \fairseq codebase to facilitate future joint research areas between speech and NLP.
\end{itemize}
\noindent We elaborate our technical rationale in the following sections.

\subsection{Input format and dataset classes}
Our speech data follows the format in Kaldi, where utterances are stored in the Kaldi-defined SCP format, consisting of space-delimited lines that follows this template:
\begin{displayquote}
  \verb|<UttID> <FeatureFile>:<Offset>|
\end{displayquote}
\noindent where \texttt{<UttID>} is the utterance ID, a key that points to any utterance in the dataset, and \texttt{<FeatureFile>} is a string interpreted as an extended filename for reading from a binary file (ARK format) storing the actual acoustic feature data, following the practice\footnote{~\url{https://kaldi-asr.org/doc/io.html}.} in Kaldi.

In theory, any kind of acoustic feature vectors (e.g. MFCC) can be stored in the feature file. In \espresso, we follow \espnet and employ the commonly used 80-dimensional log Mel feature with the additional pitch features (in total, 83 dimensions for each frame).

In \fairseq, there is a concept called ``datasets'', which contains a set of training samples and abstracts away details such as shuffling and bucketing. We follow this and create the following dataset classes in \espresso: 
\begin{itemize} \setlength\itemsep{0cm}
    \item \texttt{data.ScpCachedDataset}: this contains the real-valued acoustic features extracted from the speech utterance. Each training batch drawn from this dataset is a real-valued tensor of shape $[{\rm BatchSize} \times {\rm TimeFrameLength} \times {\rm FeatureDims}]$ that will be fed to the neural speech encoder (Section \ref{sec:lstm_encoder}). Since the acoustic features are large and cannot be loaded into memory all at once, we also implement sharded loading, where given the order of the incoming examples in an epoch, a bulk of features is pre-loaded once the previous bulk is consumed for training / decoding. This helps balance the file system's I/O load and the memory usage.
    \item \texttt{data.TokenTextDataset}: this contains the gold speech transcripts as text. Each training batch is a integer-valued tensor of shape $[{\rm BatchSize} \times {\rm SequenceLength}]$, where each integer in this tensor is the index of the character / subword unit in the token dictionary (see below).
    \item \texttt{data.SpeechDataset}: this is a container for the two datasets above: each sample drawn from this dataset contains two fields, \texttt{source} and \texttt{target}, that points to the speech utterance and the gold transcripts respectively.
\end{itemize}

\noindent Note that in speech recognition, the token dictionary (set of all vocabulary) is different from the common practice in \fairseq due to the additional special token \texttt{<space>}. For this reason, we do not directly use the \texttt{data.Dictionary} class from \fairseq, instead, we inherit that class and create our \texttt{data.TokenDictionary} class for this purpose, with the extra functionality of handling \texttt{<space>}.

For speech decoding purposes rather than NMT (default in \fairseq), normally the output unit for each decoding step is a \emph{subword unit} instead of a word, since it is shown that for ASR using whole words as modeling units is only possible when large amounts of training data (at least tens of thousands of hours) is available \cite{soltau2017neural,qudhkhasi2018building}. A subword unit can either be a \emph{character} or \emph{character sequence} like BPE \cite{sennrich2016neural} or a SentencePiece\footnote{~\url{https://github.com/google/sentencepiece}.} \cite{kudo2018subword}. Both are supported in \espresso and experimental results will follow.

\subsection{Output format}
\espresso supports two output format: a \emph{raw} format and a more detailed \emph{aligned results} version that helps debugging. 

The raw format just consists of space-delimited lines that follows this template:
\begin{displayquote}
  \verb|<UttID> <DecodedSequence>|
\end{displayquote}
\noindent where \texttt{<UttID>} is the original utterance ID from the SCP dataset, and the \texttt{<DecodedSequence>} is the raw output of the speech recognition system. 

The \emph{aligned results} provide an aligned sequence between the gold reference transcript and the predicted hypothesis. An example is shown below:

{\small{
\begin{verbatim}
  4k9c030b
  REF: "QUOTE AN EYE FOR AN EYE "UNQUOTE
  HYP: "QUOTE AN EYE FOR    ANY "END-QUOTE
  STP:                   D  S   S
  WER: 42.86%
\end{verbatim}
}}

Each such record consists of 5 rows: the first line is the utterance ID; \texttt{REF} and \texttt{HYP} is the reference transcript and the system output hypothsis respectively -- these two are aligned using minimal edit distance. The fourth line, \texttt{STP} (step), contains the error the system makes at each decoding step: it could be one of \texttt{S} (substitution), \texttt{I} (insertion) and \texttt{D} (deletion), corresponding to the three types of errors when evaluating the word error rate (WER) commonly used to evaluate speech recognition systems. The last line is WER calculated on this utterance. Such output format facilitates easy human inspection to the different error types made by the system, rendering debugging easier for researchers.

\subsection{Encoder-Decoder}
\espresso supports common sequence generation models and techniques arisen from the research in the ASR and NLP community. The \emph{de facto} standard model, the encoder-decoder with attention \cite{bahdanau2015neural,luong2015effective} (also successfully pioneered by \cite{chorowski2014end} in the speech community), is implemented as our \texttt{models.speech\_lstm.SpeechLSTMModel} class. Owing to the modularity and extensibility of \espresso, other models, e.g., Transformer \cite{vaswani2017attention}, can be easily integrated from the underlying \fairseq.

\vspace{0.2cm}
\noindent \textbf{\textit{CNN-LSTM Encoder}}\quad
\label{sec:lstm_encoder}
The default encoder we used is a 4-layer stacked 2-dimensional convolution (with batch normalization between layers), with kernel size $(3, 3)$ on both the time frame axis and the feature axis \cite{zhang2017very,watanabe2018espnet}. 2$\times$-downsampling is employed at layer 1 and 3, resulting in $\nicefrac{1}{4}$ time frames after convolution. The final output channel dimensionality is 128, with the 21 downsampled frequency features, hence a total of $128 \times 21 = 2688$ dimensional features for each time frame.

Then 3 layers of bidirectional LSTMs \cite{graves2005bidirectional} are stacked  upon the output channels yielded by the stacked convolution layers.

This architecture, with the various dimensionality, number of layers, and other hyperparameters customizable, is implemented in our \texttt{models.speech\_lstm.SpeechLSTMEncoder} class.

\vspace{0.2cm}
\noindent \textbf{\textit{LSTM Decoder with Attention}}\quad
\label{sec:lstm_decoder}
We use a 3-layer LSTM decoder by default, with Bahdanau attention \cite{bahdanau2015neural} on the encoded hidden states (Luong attention \cite{luong2015effective} is also implemented). We follow the architecture in the Google Neural Machine Translation (GNMT) system \cite{gnmt}, where the context vectors generated by the attention mechanism is fed to all 3 layers of the decoding LSTM. Residual connections \cite{he2016deep} are added between the decoder layers. These are implemented in the \texttt{models.speech\_lstm.SpeechLSTMDecoder} class.

\subsection{Training Strategies}


\vspace{0.2cm}
\noindent \textbf{\textit{Scheduled Sampling}}\quad
Scheduled sampling \cite{bengio2015scheduled} is supported by our system, with promising results in end-to-end speech recognition \cite{baskar2019promising}. With scheduled sampling, at each decoder step, the gold label is fed to the next step with $p$ probability, whereas the predicted token\footnote{~This is the token with the maximum posterior probability resulting from the previous LSTM decoder step. It may not necessarily be gold.} is fed with $(1-p)$ probability. In our implementation such mechanism will start at an intermediate epoch $N$ in the training process: the first few epochs are always trained with gold labels. The probability $p$ can be scheduled in the training process: later epochs might use a smaller probability to encourage the model not to rely on the gold labels.

\vspace{0.2cm}
\noindent \textbf{\textit{Label Smoothing}}\quad
Label smoothing \cite{szegedy2016rethinking} has been proposed to improve accuracy by computing the loss (i.e., cross entropy here) not with the ``hard'' (one-hot) targets from the dataset, but with a weighted mixture of these targets with some distribution \cite{mueller2019labelsmoothing}. We support three kinds of these distributions in \espresso:
\begin{itemize} \setlength\itemsep{0cm}
    \item \emph{Uniform smoothing} \cite{szegedy2016rethinking}: The target is a mixture of $(1-p)$ probability of the one-hot target and the rest of the $p$ probability mass uniformly distributed across the vocabulary set;
    \item \emph{Unigram smoothing} \cite{pereyra2017regularizing}: Mixed with a unigram language model trained on the gold transcripts;
    \item \emph{Temporal smoothing} \cite{chorowski2017towards}: Mixed with a distribution assigning probability mass to neighboring tokens in the transcript. Intuitively, this smoothing scheme helps the model recover from beam search errors: the network is more likely to make mistakes that simply skip a subword unit of the transcript.
\end{itemize}


\vspace{0.2cm}
\noindent \textbf{\textit{Model Selection via Validation}}\quad
At the end of each training epoch, we compute the WER on the validation set using greedy-search decoding without language model fusion (see Section \ref{sec:fusion}). This is different from the approach in previous frameworks such as \espnet and \fairseq, where they compute the loss function on the validation set (the gold labels are fed in) to perform model selection. We argue that our approach may be more suitable since free decoding on the validation set is a closer scenario to the final metric on test sets. Owing to efficiency concerns, we do not use full-blown language model fused beam search decoding for validation (arguably this is even better).

We employ learning rate scheduling following \fairseq: at the end of an epoch, if the metric on the validation set is not better than the previous epoch, the learning rate is reduced by a factor (e.g. $\nicefrac{1}{2}$). Empirically we found that the reduction of the learning rate will be less frequent if using WER as the validation metric as compared to the loss value on the validation set. According to \cite{specaugment}, a less frequent learning rate reduction generally leads to better performance.

\section{Language Model-Fused Decoding}
\label{sec:fusion}
It is shown in recent research that a pure sequence-to-sequence transduction model for ASR without an external language model component (which is used in traditional hybrid ASR systems) is far from satisfactory \cite{kannan2018analysis}. This is in contrast with neural machine translation (NMT) models, where normally no external language model is needed. This performance gap is hypothesized to be caused by the fact that the ASR model is only trained on speech-transcript pairs. The gold transcript set is not large enough to produce state-of-the-art neural language models, which are typically trained on a corpus on the scale of 1 billion words. 

In \espresso, we employ \emph{shallow fusion} \cite{gulcehre2015using} as a language model integration technique, which is proven to be effective in speech recognition \cite{kannan2018analysis,toshniwal2018comparison}. The LSTM decoder with shallow fusion computes a weighted sum of two posterior distributions over subword units: one from the end-to-end speech recognition model, the other from the external neural language model. 

We support 3 types of external neural language models:
\begin{itemize} \setlength\itemsep{0cm}
    \item \emph{Subword-unit language model}: A vanilla LSTM-based language model trained on subword units. Here subword units can either just be characters (with \texttt{<space>} as an additional special token) or trained subword units (e.g. BPE \cite{sennrich2016neural} or SentencePiece \cite{kudo2018subword});
    \item \emph{Multi-level language model} \cite{hori2017multilevel}: This is a combination of character-based and word-based language models. Hypotheses in the beam are first scored with the character-based language model until a word boundary (\texttt{<space>}) is encountered. Known words are then re-scored using the word-based language model, while the character-based language model provides for out-of-vocabulary scores;
    \item \emph{Look-ahead word-based language model} \cite{hori2018endtoend}: This model enables outputting characters for each decoding step with a pre-trained word-based language model, by providing look-ahead word probabilities based on the word prefix (sequence of characters) decoded. This is shown in \cite{hori2018endtoend} to be superior to the multi-level language model.
\end{itemize}

\subsection{Parallelization with Look-ahead Word-based LMs}
\label{sec:lookahead_lm}
The original implementation of the look-ahead word-based language model in \espnet \cite{watanabe2018espnet} is not operating on batches, making the decoding speed slow. In \espresso, we devise a fully-parallelized version of the decoding algorithm on GPUs.

\begin{figure}[t]
    \centering
    \includegraphics[width=6cm]{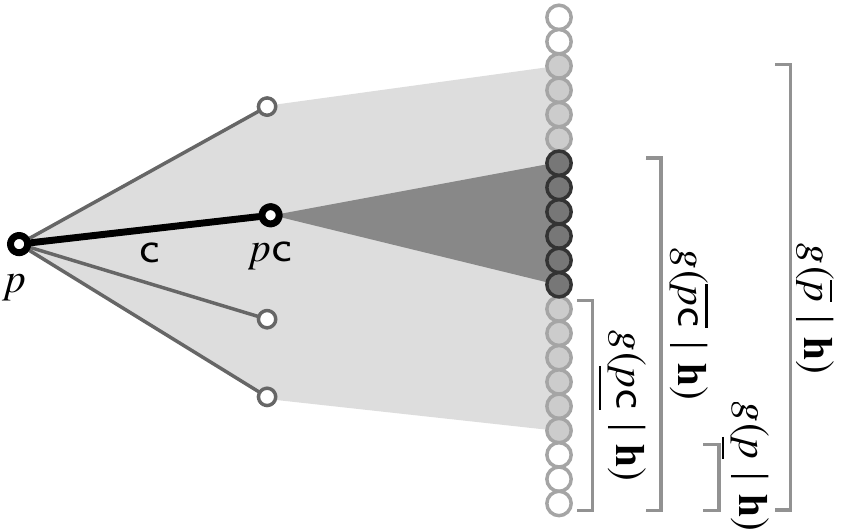}
    \caption{The efficient look-ahead LM algorithm. The dark gray and light gray subtrees correspond to the probability mass spanned by prefix $p\ch$ and $p$ (numerator and denominator in Eq. (\ref{eq:lookahead-lm})) respectively. These can be efficiently computed via Eq. (\ref{eq:fast-lm}) using \texttt{cumsum}.}
    \label{fig:prefix-tree}
\end{figure}

In \cite{hori2018endtoend}, a word-based language model is converted to a character-based one via a technique using \emph{prefix trees}. The \emph{prefix tree automaton} $T = (\Sigma, Q, \varepsilon, \tau, V)$ is a finite-state automaton (see Fig. \ref{fig:prefix-tree}):
\begin{itemize}
    \item $\Sigma$ is the character set (including \texttt{<space>});
    \item $V \subseteq \Sigma^*$ is the \emph{word vocabulary set} and also the final state set; 
    \item $Q = \{\prefix \sqsubseteq \word \mid \word \in V \} \subseteq \Sigma^*$ is the set of all \emph{prefixes}\footnote{~We 
denote ``$\prefix$ is a prefix of $\prefixq$'' as $\prefix \sqsubseteq \prefixq$: e.g. ``stand'' $\sqsubseteq$ ``standard''. } of the words in $V$ and also the state set; 
    \item $\varepsilon$ is the empty string, which also serves as the initial state;
    \item $\tau: Q \times \Sigma \to Q$ is the state transition function, where given a state and an input character, $\tau(\prefix, \ch) = \prefix \ch$, i.e. a simple concatenation. 
\end{itemize}

A look-ahead word-based LM computes the probability of the next character $\ch \in \Sigma$ based on a given word history $\history$ and a word prefix $\prefix \in Q$ (i.e., a state in the prefix tree automaton): 
\begin{equation}
\label{eq:lookahead-lm}
    P(\ch \mid \prefix, \history) = \dfrac{{\displaystyle\sum}_{\suffix: ~ \prefix \ch \suffix \in V} ~ P_{\rm W}(\prefix \ch \suffix \mid \history) } { {\displaystyle\sum}_{\suffix: ~\prefix\suffix \in V} ~ {P_{\rm W}(\prefix\suffix \mid \history )}} \ .
\end{equation}

\noindent where $P_{\rm W}(\word \mid \history)$ is the probability of the word $\word$ predicted by the word-based LSTM language model. In Eq. (\ref{eq:lookahead-lm}), the numerator is the sum of the probability of all words prefixed by $p\ch$, i.e. all possible words that could be generated from $p\ch$ if the state is moved from $p$ to $p\ch$; the denominator is the sum of the probability of all possible words at the current state $p$ (see Fig. \ref{fig:prefix-tree}).

\cite{hori2018endtoend} proposed an efficient way to compute the sum in Eq.~(\ref{eq:lookahead-lm}). We denote $\prefix$ precedes $\prefixq$ \emph{lexicographically} as $\prefix \prec \prefixq$, and define the \emph{upper bound} $\overline{\prefix}$ (the lexicographically greatest element prefixed by $\prefix$) and \emph{lower bound} $\underline{\prefix}$ (the greatest element lexicographically less than any word prefixed by $\prefix$) as:
\begin{equation}
  \overline{\prefix} = \max_{\word \in V, \prefix \sqsubseteq \word} \word; \quad \underline{\prefix} = \max_{\word \in V, \prefix \not\sqsubseteq \word, \word \prec \prefix} \word
\end{equation}

Given that the vocabulary set is sorted lexicographically, we can efficiently compute the sum of the probability of all words \emph{preceding or equal to} a given word, using efficient routines like \texttt{cumsum}:
\begin{equation}
    g(\word \mid \history) = \sum_{\word^\prime \preceq \word} P_{\rm W}(\word^\prime \mid \history)
\end{equation}

Using these definitions, Eq.~(\ref{eq:lookahead-lm}) can be rewritten as
\begin{equation}
\label{eq:fast-lm}
    P(\ch \mid \prefix, \history) = \dfrac{g(\overline{\prefix \ch} \mid \history) - g(\underline{\prefix \ch} \mid \history)}{g(\overline{\prefix} \mid \history) - g(\underline{\prefix} \mid \history)}    
\end{equation}

This method is illustrated in Fig.~\ref{fig:prefix-tree}, showing that the probability of a character given a prefix can be efficiently computed via a \texttt{cumsum} operation and simple arithmetics.

In our parallelized implementation, each prefix $p$ (corresponding to a state in the automaton) is indexed as a unique integer. Hence the batch of decoding states is compactly stored as a vector of integers, each corresponding to a state on the prefix tree automaton. The automaton itself is stored as a matrix $\bf T$ with shape $[{\rm NumberOfStates}\times{\rm MaxOutDegree}]$, where the row $T_\prefix$ contains the index of all descendants of $\prefix$, logically forming an adjacency list. The index of the $\overline{\prefix}$ and $\underline{\prefix}$ for each state $\prefix$ is also precomputed and cached. In sum, the entire prefix tree automaton is vectorized.

To compute $P(\ch \mid \prefix, \history)$ for all $\ch \in \Sigma$ over batches and beam hypotheses, the following steps are executed:
\begin{enumerate}[label=(\roman*)]
    \item Update $P_{\rm W}(\word \mid \history)$ for all $\word \in V$ from a specific decoding step of the word-based language model for those hypotheses that encounter the end-of-word (\texttt{<space>}) symbol in the batch;
    \item Update the $g(\word \mid \history)$ function using $P_{\rm W}(\word \mid \history)$ for all $\word \in V$;
    \item Get all possible successive states;
    \item Get all upper and lower bounds for all successive states;
    \item Compute the probability for each $\ch \in \Sigma$ according to Eq.~(\ref{eq:fast-lm}).
\end{enumerate}
Note that the first step follows a batched forward computation of a neural language model; the third and the fourth steps can be computed via tensorized advanced indexing; and the second and the last steps can be executed using vectorized arithmetics. Hence we obtain a fully parallelized algorithm that runs on GPUs.

As mentioned in the first step, at a specific decoding step, some elements in the batch may encounter the end-of-word symbol, whereas others may not: running this conditional operation requires special treatment. We devise an algorithm that shares the spirit with \cite{ding-koehn-2019-parallelizable}, an parallelized algorithm for stack LSTM parsing: first we record the elements in the batch that has not reached the end of a word with a mask, then progress the state for one step for all of these elements, finally, for those masked states, their original states are restored, and only $P_{\rm W}(\word \mid \history)$ of those not masked are updated.

\subsection{Improved Coverage}
With language model fusion, the decoder tends to make more deletion errors when the language model weight becomes larger \cite{bahdanau2016end}. A coverage term (a scalar value assigned to each hypothesis in the beam) is first proposed in \cite{chorowski2017towards} to promote longer transcripts and also to prevent attentions from looping over utterance (repeating certain $n$-grams when decoding) when using shallow fusion:
\begin{eqnarray}
\label{eqn:coverage_original}
\textrm{coverage} = {\displaystyle\sum}_j~ \mathbbm{1}\left[{\displaystyle\sum}_i~ a_{ij} > \tau\right]
\end{eqnarray}
where $a_{ij}$ is the attention weight for the decoder step $i$ and encoder frame $j$, $\tau>0$ is a tunable hyper-parameter, and $\mathbbm{1}[\cdot]$ is the indicator function. This is the total number of encoder frames that has been sufficiently ``attended'' to. However, based on our experiments, applying the coverage term Eq.~(\ref{eqn:coverage_original}) is not sufficient to prevent words from being repeated. Instead we propose a modified version of the coverage term which penalizes looping attentions more aggressively:
\begin{eqnarray}
\label{eqn:coverage}
\textrm{coverage} &=& {\displaystyle\sum}_j \left(\mathbbm{1}\left[{\displaystyle\sum}_i~ a_{ij} > \tau_1\right] \right.\nonumber \\
&&\left.-\mathbbm{1}\left[{\displaystyle\sum}_i~ a_{ij} > \tau_2\right] \cdot \left( c + {\displaystyle\sum}_i~ a_{ij} - \tau_2 \right) \right)
\end{eqnarray}

\noindent where $\tau_2>\tau_1>0,c>0$ are tunable hyper-parameters. While the first term in Eq.~(\ref{eqn:coverage}) is exactly the same as Eq.~(\ref{eqn:coverage_original}), the second term penalizes the hypothesis score when the accumulated attention on encoder frame $j$ exceeds $\tau_2$. Specifically, if the accumulated attention weight on frame $j$ exceeds $\tau_1$ but not $\tau_2$, only the first term is activated, increasing the coverage score to encourage more attention on frame $j$; if the accumulated attention weight further exceeds $\tau_2$, the second term (with the minus sign) is also activated and its magnitude is the amount of the exceeding value plus a constant $c$, discouraging further attention accumulated on the same frame. Therefore, the new coverage mechanism enforces a soft constraint on the accumulated attention weight on each frame to be between $\tau_1$ and $\tau_2$, leading to both less deletion errors and less repeating $n$-grams (shown in the WSJ part of Section \ref{sec:res}). During beam search decoding this new coverage term as a whole is added to the hypothesis score with a weight (e.g. 0.01). In our experiments, $\tau_1=0.5,\tau_2=1.0,c=0.7$.

\subsection{EOS Threshold}
We implement the end-of-sentence threshold technique proposed in \cite{hannun2019sequence} to bias the decoder away from short transcriptions when decoding with a fused language model. End-of-sentence (\texttt{<eos>}) tokens can only be emitted if its probability is greater than a specific factor of the top output token candidate during beam search:
\begin{equation}
    \log P(\texttt{<eos>} \mid \textbf{h}) > \gamma \cdot \max_{t \in V} \log P(t \mid \textbf{h})
\end{equation}
where $V$ is the vocabulary set. In our experiments, $\gamma$ is set to 1.5.

\section{Recipes and Results}
\label{sec:res}
\espresso provides running recipes for a variety of well-known data sets. We elaborate the details of our recipes on Wall Street Journal \cite{paulb92} (WSJ), an 80-hour English newspaper speech corpus, LibriSpeech \cite{panayotovcpk15}, a corpus of approximately 1,000 hours of read English speech, and Switchboard \cite{godfrey1992switchboard} (SWBD), a 300-hour English telephone speech corpus. 

Besides the transcripts, all of these data sets have their own extra text corpus for training language models. Input and output word embeddings are tied \cite{press2017using} to reduce model size. 
All the models are optimized using Adam \cite{kingma2015adam} with an initial learning rate $10^{-3}$, and then halved if the metric on the validation set at the end of an epoch does not improve over the previous epoch. The training process stops if the learning rate is less than $10^{-5}$. Curriculum learning \cite{bengio2009curriculum}, which is quite helpful to stabilize training with long sequences (e.g. LibriSpeech) and improve performance (esp. SWBD), is employed for the first 1 (LibriSpeech) or 2 (WSJ / SWBD) epochs. All the models are trained / evaluated using NVIDIA GeForce GTX 1080 Ti GPUs. If not otherwise specified, all the models throughout this paper are trained with 2 GPUs using \fairseq built-in distributed data parallellism. Note that no data augmentation techniques such as speed-perturbation \cite{ko2015audio} or the more recent SpecAugment \cite{specaugment} is applied.

The hyper-parameters for the recipes are listed in \autoref{tab:hyperparameters}. 
\begin{table}[tb]
    \centering
    \caption{Hyper-parameters for the three recipes.}
    \begin{adjustbox}{max width=\linewidth}
    \begin{tabular}{rcccccc}
        \toprule
        \multirow{2}{*}{Hyper-parameter} & \multicolumn{2}{c}{WSJ} & \multicolumn{2}{c}{LibriSpeech} & \multicolumn{2}{c}{SWBD} \\
        \cmidrule(lr){2-3} \cmidrule(lr){4-5} \cmidrule(lr){6-7}
        & LM & ASR & LM & ASR & LM & ASR \\
        \midrule
        Vocab. size  & 65k & 52 & 5k & 5k & 1k & 1k \\
        Encoder \# layers & - & 3 & - & 4 & - & 4 \\
        Decoder \# layers & 3 & 3 & 4 & 3 & 3 & 3 \\
        Emb. dim.  & 1,200  & 48 & 800 & 1,024 & 1,800 & 640\\
        Hidden dim. & 1,200  & 320 & 800 & 1,024 & 1,800 & 640\\
        \# Params. & 113M & 18M & 25M & 174M & 80M & 70M\\
        Dropout rate     & 0.35   & 0.4 & 0.0 & 0.3 & 0.3 & 0.5\\
        Avg. batch size & 435 & 48 & 1,733 & 34 & 1,783 & 69\\
        \midrule
        Beam size & \multicolumn{2}{c}{50} & \multicolumn{2}{c}{60} & \multicolumn{2}{c}{35} \\
        LM fusion weight & \multicolumn{2}{c}{0.9} & \multicolumn{2}{c}{0.47} & \multicolumn{2}{c}{0.25} \\
        \bottomrule
    \end{tabular}
    \end{adjustbox}
    \label{tab:hyperparameters}
\end{table}


\vspace{0.2cm}
\noindent \textbf{\textit{WSJ}}\quad
We adopt the look-ahead word-based language model \cite{hori2018endtoend} as the external language model. We report the perplexities on the validation / evaluation set: the external word-based language model achieves 72.0 perplexity on dev93 and 59.0 on eval92. 

For the encoder-decoder model, the vocabulary size of subword units (characters) for WSJ is 52, the same as in \espnet.\footnote{~It includes 45 characters constituting the training transcripts, plus 3 atomic symbols: \texttt{<*IN*>}, \texttt{<*MR.*>} and \texttt{<NOISE>}, plus 4 special symbols: \texttt{<pad>}, \texttt{<eos>}, \texttt{<unk>} and \texttt{<space>}.} 
Temporal label smoothing with $p=0.05$ and scheduled sampling with $p=0.5$ starting at epoch 6 are adopted. 

Baseline end-to-end systems are compared: Hadian et al. \cite{hadian2018end}, an end-to-end\footnote{~This is a hybrid system. ``End-to-end'' here is in the sense that it does not need HMM-GMM training or tree-building steps.} model with the lattice-free MMI objective \cite{povey2016purely}; Baskar et al. \cite{baskar2019promising}, an encoder-decoder model with discriminative training in \espnet; Zeghidour et al. \cite{zeghidour2018fully}, a pure convolutional network with ASG loss \cite{collobert2016wav2Letter}; Likhomanenko et al. \cite{Likhomanenko2019who}, a lexicon-free decoding method with the acoustic model proposed in \cite{zeghidour2018fully}; and the last one, Deep Speech 2 \cite{amodei2016deep}.

We show the beam search decoding results of various configurations of \espresso in Table~\ref{tab:wsj_lstm} with beam size 50. The breakdown of the three kinds of errors is shown in Table~\ref{tab:wsj_lstm_breakdown}. The first row gives WERs where no language model fusion is applied. The second row is after integrating the look-ahead word-based language model, with its optimal LM fusion weight 0.5. Although it has already significantly reduced the overall WER, deletion errors increase. Further applying the improved coverage yields better performance by suppressing deletion errors. 
If we only use the first term in Eq.~(\ref{eqn:coverage}) which is equivalent to the coverage term in \cite{chorowski2017towards}, the insertions errors will increase from 0.8 to 1.3 on dev93, and from 0.6 to 0.9 on eval92. A manual inspection of the decoded results reveals that these additional insertions are mostly repeated words. This observation validates the effectiveness of our improved coverage mechanism. Alternatively, if we apply the EOS threshold, we achieve state-of-the-art performance on WSJ among end-to-end models.

\begin{table}[tb]
  \caption{WERs (\%) on the WSJ dev93 and eval92 set.}
  \label{tab:wsj_lstm}
  \centering
  \begin{adjustbox}{max width=\linewidth}
  \begin{tabular}{l c c}
    \toprule
    & dev93 &  eval92 \\
    \midrule 
    Hadian et al.\tablefootnote{~The result is with ``full bichar'' (using all possible 2-gram characters as context-dependent modeling units), data speed-perturbation, and a 3-gram word language model.} \cite{hadian2018end} & - & 4.1 \\
    Baskar et al. (\espnet) \cite{baskar2019promising} & - & 3.8 \\
    Likhomanenko et al. \cite{Likhomanenko2019who} & 6.4 & 3.6 \\
    Zeghidour et al. \cite{zeghidour2018fully} & 6.8 & 3.5 \\
    Amodei et al.\tablefootnote{~It uses 12,000 hours transcribed data for acoustic model training and large crawled text for language model training, making it not comparable.} (Deep Speech 2) \cite{amodei2016deep} & \it\textcolor{gray}{4.4} & \it\textcolor{gray}{3.1} \\
    \midrule
    \espresso LSTM & 14.8 & 12.1 \\
    \quad+Look-ahead Word LM & 7.4 & 5.1  \\
    \quad\quad+Improved Coverage & \bf 5.9 & 3.5 \\ 
    \quad\quad+EOS Threshold & \textbf{5.9} & \textbf{3.4} \\
    \bottomrule
  \end{tabular}
  \end{adjustbox}
\end{table}

\begin{table}[tb]
  \caption{Breakdown of the WERs (\%) on WSJ.}
  \label{tab:wsj_lstm_breakdown}
  \centering
  \small
  \begin{adjustbox}{max width=\linewidth}
  \begin{tabular}{lcccccc}
    \toprule
    & \multicolumn{3}{c}{dev93} & \multicolumn{3}{c}{eval92} \\
    \cmidrule(lr){2-4} \cmidrule(lr){5-7}
    & Sub & Ins & Del &  Sub & Ins & Del \\
    \midrule
    \espresso LSTM & 12.0 & 1.4 & 1.4 & 9.7 & 1.5 & 1.0 \\
    \quad+Look-ahead Word LM & 4.6 & \bf 0.8 & 2.0 & 3.1 & 0.7 & 1.3 \\
    \quad\quad+Improved Coverage & 4.3 & \bf 0.8 & \bf 0.8 & 2.7 & 0.6 & \bf 0.3 \\
    \quad\quad+EOS Threshold & \bf 4.1 & 0.9 & \bf 0.8 & \bf 2.6 & \bf 0.5 & \bf 0.3 \\
    \bottomrule
  \end{tabular}
  \end{adjustbox}
\end{table}

\vspace{0.2cm}
\noindent \textbf{\textit{LibriSpeech}}\quad
SentencePiece is used as the subword units in our LibriSpeech setup for both external language modeling and encoder-decoder model. 
We combine dev-clean and dev-other sets together as a single validation set for both language model and encoder-decoder model training.


Again, we report the perplexities on the validation / evaluation sets. \espresso obtains 35.4 and 37.3 on the dev-clean and dev-other sets, and 37.2 and 37.2 on test-clean and test-other.

Uniform smoothing with $p=0.1$ is applied throughout the entire training. No scheduled sampling is used. The vanilla shallow fusion is used without the ``look-ahead'' technique.  The results, along with several baseline systems, are demonstrated and compared in Table~\ref{tab:librispeech_lstm}, 

\begin{table}[tb]
  \caption{WERs (\%) on the LibriSpeech dev and test sets.}
  \label{tab:librispeech_lstm}
  \centering
  \setlength\tabcolsep{3pt}
  \begin{adjustbox}{max width=\linewidth}
  \begin{tabular}{l c c c c}
    \toprule
    
    & \multicolumn{2}{c}{dev} & \multicolumn{2}{c}{test} \\
    \cmidrule(lr){2-3} \cmidrule(lr){4-5}
    & clean &  other & clean & other \\
    \midrule
    Zeghidour et al. \cite{zeghidour2018fully} & 3.1 & 10.0 & 3.3 & 10.5 \\
    Hannun at al. \cite{hannun2019sequence} & 3.0 & 8.9 & 3.3 & 9.8 \\
    Park et al. \cite{specaugment} (w/o SpecAugment) & - & - & 3.2 & 9.8 \\
    L{\"{u}}scher et al. \cite{luscher2019hybrid} & \bf 2.6 & \bf 8.4 & \bf 2.8 & 9.3 \\
    \midrule
    \espresso LSTM & 3.8 & 11.5 & 4.0 & 12.0 \\
    \quad+Subword LM & 3.3 & 8.9 & 3.4 & 9.5 \\
    \quad\quad+Improved Coverage & 2.9 & 8.8 & 3.2 & 9.0 \\
    \quad\quad+EOS Threshold & 2.8 & \bf 8.4 & \bf 2.8 & \bf 8.7 \\
    \bottomrule
  \end{tabular}
  \end{adjustbox}
\end{table}

We can see that both the improved coverage or EOS threshold help in this setup as well, where actually deletion error reductions contribute mostly. In addition, we achieve state-of-the-art results on end-to-end models for LibriSpeech without any data augmentation.

\vspace{0.2cm}
\noindent \textbf{\textit{Switchboard}}\quad
The vocabulary consists of 1,000 subword units segmented by SentencePiece\footnote{~It includes additional special tokens \texttt{[laughter]}, \texttt{[noise]}, and \texttt{[vocalized-noise]}.} trained on both Switchboard and Fisher transcripts. As there is no official validation set, we hold out the same 4,000-example subset of the training data as in Kaldi for validation. 
We apply scheduled sampling starting at epoch 6 with probability from 0.9 to 0.6. Uniform smoothing is used with $p = 0.1$.

The language model achieves a validation perplexity of 17.3. 
No coverage is used during decoding. The results of our current system and 3 other competitive end-to-end baselines are shown in Table~\ref{tab:swbd_lstm}. Again, we obtain state-of-the-art results without SpecAugment. The coverage term or EOS threshold does not help on this dataset, and we suspect it is because its optimal LM fusion weight is not as large as those for the other two datasets, resulting in less deletion errors.

\begin{table}[tb]
  \caption{WERs (\%) on the SWBD Hub5'00 evaluation set.}
  \label{tab:swbd_lstm}
  \centering
  \begin{adjustbox}{max width=\linewidth}
  \begin{tabular}{l c c}
    \toprule
    & Switchboard & CallHome \\
    \midrule
    Cui et al. \cite{cui2018improving} (w/ speed-pertubation) & 12.0 & 23.1 \\
    Zeyer et al. \cite{zeyer2018comprehensive} & 11.0 & 23.1 \\
    Park et al. \cite{specaugment} (w/o SpecAugment) & 10.9 & 19.4 \\
    \midrule
    \espresso LSTM & 10.7 & 20.7 \\
    \quad+Subword LM & \textbf{9.2} & \textbf{19.1} \\
    \bottomrule
  \end{tabular}
  \end{adjustbox}
\end{table}

\section{Training and Decoding Speed}

In this section we compare \espresso and \espnet on the training and decoding time with single GPU, using the WSJ dataset. 

For fair comparison, we create architectures in \espresso (different from those in Section \ref{sec:res}) that mimics the WSJ recipe in \espnet as closely as possible. Data preparation and vocabulary building are identical. The neural architecture is mostly the same (e.g. number and dimension of LSTM layers), with a few minor exceptions: e.g. \espnet's use of location-based attention (which \espresso does not employ),  pooling CNN layers (\espresso uses strided CNN without pooling), and joint cross-entropy+CTC loss (\espresso uses only cross-entropy loss).

Table~\ref{tab:train_speed} gives training wall time comparisons on both the external word-based language model and the encoder-decoder model, which are averaged over 20 epochs and 15 epochs respectively. \espresso is 17.9\% faster than \espnet on the language model training, and 16.0\% faster on the encoder-decoder model training. We conjecture that the main reason of such speed gain for language model training is that in \fairseq (and hence in \espresso) training examples are sorted based on input sequence lengths before batching (i.e., bucketing; \espnet does not use it for language modeling), so that the average sequence length in batches is smaller.

\begin{table}[tb]
  \caption{Training (per epoch) and decoding wall time on WSJ.}
  \label{tab:train_speed}
  \centering
  \begin{adjustbox}{max width=\linewidth}
  \begin{tabular}{c c c c c}
    \toprule
    & \multicolumn{2}{c}{Training} & \multicolumn{2}{c}{ASR Decoding (eval92)} \\
    \cmidrule(lr){2-3} \cmidrule(lr){4-5} 
    & LM & ASR & w/o LM & w/ look-ahead LM \\
    \midrule
    \espnet & 56min & 36min & 5min 21s & 29min 16s \\ 
    \espresso & \textbf{46min} & \textbf{31min} & \bf 1min 27s & \bf 2min 44s \\
    \midrule
    Speedup & 17.9\% & 16.0\% & 3.7$\times$ & 10.7$\times$ \\
    \bottomrule
  \end{tabular}
  \end{adjustbox}
\end{table}

A notable advantage of \espresso compared to \espnet is the decoding speed. In order to have a more fair comparison, we enable GPU batch decoding in \espnet \cite{seki2018vectorization}, and make batch and beam sizes of the two systems the same. We measure the decoding time on the WSJ eval92 data set, which consists of 333 utterances. Table~\ref{tab:train_speed} shows that, without language model fusion, \espresso is 3.7$\times$ faster than \espnet. With the look-ahead language model fusion, the speedup is even more prominent---more than 10$\times$ faster---mostly due to our parallelized implementation of the look-ahead language model fusion (cf. Section \ref{sec:lookahead_lm}), as opposed to \espnet, where LM scores are computed iteratively over examples within a minibatch.


\section{Conclusion}
\label{sec:print}
In this paper we present \espresso, an open-source end-to-end ASR toolkit built on top of \fairseq, an extensible neural machine translation toolkit. In addition to advantages inherited from \fairseq, \espresso supports various other features for ASR that are seamlessly integrated with \fairseq, including reading data in Kaldi format, and efficient parallelized language model-fused decoding. We also provide ASR recipes for WSJ, LibriSpeech, and Switchboard data sets, and achieve state-of-the-art performance among end-to-end systems. By sharing the underlying infrastructure with \fairseq, we hope \espresso will facilitate future joint research in speech and natural language processing, especially in sequence transduction tasks such as speech translation and speech synthesis. 

\bibliographystyle{IEEEbib}
\bibliography{refs}

\end{document}